%% file: main.tex
\documentclass[final,5p,times,twocolumn]{elsarticle}

\usepackage{lipsum}
\usepackage{multirow}
\usepackage{makecell}
\usepackage{array}
\usepackage{booktabs}
\usepackage{amssymb}
\usepackage{amsthm,amsmath}
\usepackage{mathrsfs}
\usepackage{xcolor}

\journal{Pattern Recognition Letters}

\begin{document}

\def\eg{\emph{e.g.},}
\def\etal{\emph{et al.}}
\def\ie{\emph{i.e.},}
\def\etc{\emph{etc.}}
\def\wrt{w.r.t.}
\def\vs{\emph{vs.}}
\def\blue#1{\textcolor{blue}{#1}}
\def\red#1{\textcolor{red}{#1}}
\def\green#1{\textcolor{green}{#1}}

\begin{frontmatter}



\title{Learning A Robust RGB-Thermal Detector for Extreme Modality Imbalance} 


\author[lab_1]{Chao Tian}
\ead{tianchao@stu.hit.edu.cn}
\author[lab_1]{Chao Yang}
\ead{20b951014@stu.hit.edu.cn}
\author[lab_1]{Guoqing Zhu}
\ead{20B951002@stu.edu.hit.cn}
\author[lab_1]{Qiang Wang}
\ead{qiang.wang@hit.edu.cn}
\author[lab_1]{Zhenyu He\corref{corr_author}}
\ead{zhenyuhe@hit.edu.cn}

\affiliation[lab_1]{organization={School of Computer Science and Technology, Harbin Institute of Technology},
            city={Shenzhen},
            state={Guangdong},
            country={China}}

\cortext[corr_author]{Corresponding author}

\input{0_abstract}


\begin{keyword}
RGB-Thermal, Object Detection, Modality Imbalance
\end{keyword}

\end{frontmatter}

\input{1_intro}

\input{2_related}

\input{3_method}

\input{4_experiment}

\input{5_conclu}

\bibliographystyle{elsarticle-num-names}
\bibliography{main}

\end{document}

%% file: 0_abstract.tex
\begin{abstract}

RGB-Thermal (RGB-T) object detection utilizes thermal infrared (TIR) images to complement RGB data, improving robustness in challenging conditions. Traditional RGB-T detectors assume balanced training data, where both modalities contribute equally. However, in real-world scenarios, modality degradation—due to environmental factors or technical issues—can lead to extreme modality imbalance, causing out-of-distribution (OOD) issues during testing and disrupting model convergence during training. This paper addresses these challenges by proposing a novel base-and-auxiliary detector architecture. We introduce a modality interaction module to adaptively weigh modalities based on their quality and handle imbalanced samples effectively. Additionally, we leverage modality pseudo-degradation to simulate real-world imbalances in training data. The base detector, trained on high-quality pairs, provides a consistency constraint for the auxiliary detector, which receives degraded samples. This framework enhances model robustness, ensuring reliable performance even under severe modality degradation. Experimental results demonstrate the effectiveness of our method in handling extreme modality imbalances~(decreasing the Missing Rate by 55\%) and improving performance across various baseline detectors.

\end{abstract}

%% file: 1_intro.tex
\section{Introduction}
\label{sec:intro}

\begin{figure}[!t]
\centering
\includegraphics[width=\linewidth]{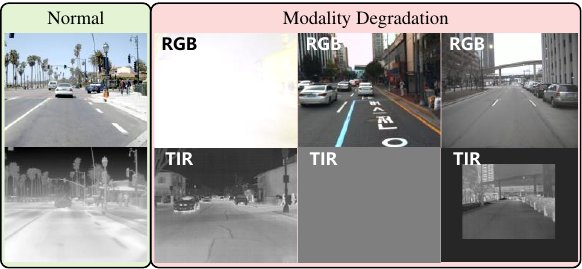}
\caption{Illustration of the modality degradation. The modality degradation would occur due to electrical failure or other unexpected reasons. 
The degradation causes the out-of-distribution issue in testing and disturbs the convergence in training.
}
\label{Fig:intro}
\vspace{-10pt}
\end{figure}

Object detection plays a crucial role in computer vision tasks, particularly in autonomous driving~\cite{zhang2020multispectral_flir, kaist, bijelic2020seeing} and security surveillance~\cite{bijelic2020seeing, fuhr2015camera}, where robust detection systems are critical. Traditional object detection methods relying solely on RGB images\cite{abnet} often struggle in challenging environmental conditions, such as fog, low-light, or inclement weather~\cite{ku2018joint}. Thermal infrared (TIR) imagery stands out due to its insensitivity to weather and lighting conditions~\cite{liu2016multispectral, tian2023cross}. Consequently,
many works~\cite{zhang2019weakly, gaff, li2022confidence, chen2022multimodal, lee2024crossformer, shao2024mod} resolve to thermal infrared (TIR) images to complement RGB images,~\ie RGB-T object detection.
Recent advancements~\cite{liu2016multispectral, zhang2019weakly, gaff, kim2021uncertainty, mbnet} in RGB-T object detection assume that both RGB and TIR modalities are available during both the training and testing phases. However, in real-world scenarios, cameras may experience complete modality degradation due to environmental challenges or technical malfunctions, leading to the loss of one modality entirely. Additionally, due to differences in field-of-view (FOV) between RGB and infrared cameras, some areas of the image may be captured by only one modality after registration. These issues, modality degradation on both global and local scales, can lead to extreme out-of-distribution (OOD) modality imbalance during testing, as illustrated in Figure~\ref{Fig:intro}.

\begin{figure*}[!ht]
\centering
\includegraphics[width=0.7\linewidth]{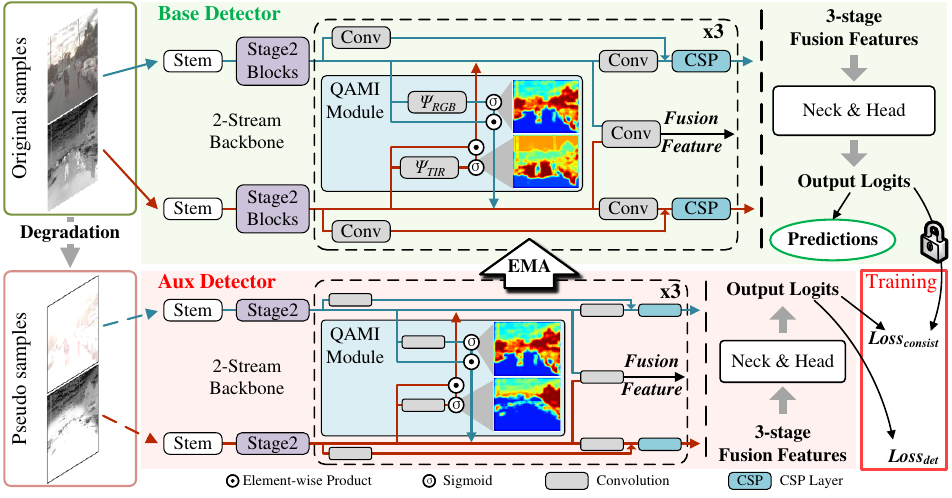}
\caption{Illustration of our proposed architecture. 
The green area is the base detector, where the detailed architecture of the backbone is presented. The proposed interaction module evaluates the quality of each modality and reweights them before fusion.
The auxiliary detector~(red) has the same network architecture as the base detector and updates by supervised training. The balanced original samples are fed to the base detector while the degraded ones~(maybe the original ones) are fed to the auxiliary detector. During supervised training, aside from the detection loss of the auxiliary detector, the consistency of logits between the base and auxiliary detector is constrained. The stem means the fore-modules of each stream and the CSP layer is a standard module in the CSPDarknet~\cite{yolov4}.
}
\label{Fig:method}
\vspace{-10pt}
\end{figure*}

Our approach introduces a novel base-and-auxiliary detector architecture, which enables the model to learn from both balanced and imbalanced RGB-T image pairs, ensuring robustness in both the training and testing phases. 
We simulate two versions of the RGB-T image pair: the original and a degraded version. They are respectively fed into two unshared networks~(\ie the base and auxiliary one) during training, and their responses are constrained by a Euclidean distance loss to ensure consistency across both versions. 
The base detector’s parameters are updated via the exponential moving average~(EMA) manner, while the auxiliary detector is supervised with end-to-end training. This setup ensures that the base detector learns from both high-quality and degraded samples, which makes it robust against modality imbalance in testing. 
After training, the base detector, which contains comprehensive knowledge for handling different conditions, is used for final testing.
The main contributions of our work are as follows:
\begin{itemize}
\item We propose a base-and-auxiliary detector architecture that effectively handles both balanced and imbalanced RGB-T image pairs, enhancing convergence and robustness.
\item We design a modality interaction module for adaptive feature fusion, resulting in a modality-quality-aware two-stream detector for RGB-T data.
\item Extensive experiments demonstrate the effectiveness of our method in handling extreme modality imbalance and improving the performance of various baseline detectors.
\end{itemize}

%% file: 2_related.tex
\section{Related Work}
\label{sec:related_work}
\noindent \textbf{RGB-T object detection.}
RGB-T object detection achieves great development with the growing datasets~\cite{kaist, cvc14, llvip, zhang2020multispectral_flir}.
The RGB-T object detection utilizes different modalities to complement each other~\cite{guo2024contrastive}. So the methods can be divided by the stage of modality fusion~\cite{chen2022multimodal}, namely image fusion, feature fusion, and box fusion. 
The image fusion-based method~\cite{wagner2016multispectral} extracts features directly from a channel-wise cascaded image and predicts like an RGB detector. It suffers from insufficient feature extraction and has become obsolete nowadays. Meanwhile, box fusion-based methods~\cite{li2022confidence,chen2022multimodal} leverage the theory of probabilistic ensemble to integrate the boxes predicted by two modalities separately. These approaches independently deploy three sub-networks for RGB, TIR, and fusion modality, a time-consuming process that is not widely adopted.
Feature fusion-based methods~\cite{wagner2016multispectral, cian} conduct feature fusion more efficiently and avoid the redundant results prediction, which are paid more attention nowadays.
In feature fusion, RGB and TIR images contribute unequally, whereas the low-quality one may introduce noise. To deal with this issue, RGB-T detectors deploy a module to evaluate the quality of each modality and re-weight them. Two-stage detectors~\cite{kim2021uncertainty, li2019illumination, wanchaitanawong2021multi} re-weight the region of interest~(RoI) features extracted within proposals. Some of the single-stage detectors~\cite{li2022confidence, kim2021mlpd, mbnet, ni2022modality} apply the modal weights on the global scale while the other work~\cite{gaff} applies the spatial-aware weights by masks.

\noindent \textbf{Learning for imbalanced data.}
Most RGB-T detectors mentioned above consider the imbalanced data identically in the testing and training stage, ignoring the degradation brought by different distributions in harsh conditions in the testing stage. 
In multimodal learning, there has been some research on handling missing modalities in the training~\cite{shi2020relating} and testing~\cite{tsai2018learning} stage separately, or in both of them~\cite{ma2021smil}. 
In multimodal segmentation, complementing the missing modality~\cite{yu20183d, sharma2019missing} is the intuitive solution. Some works~\cite{azad2022smu} resort to the distillation framework, to keep the consistency of features in modality missing. And cross-modality learning~\cite{zhang2023illumination}, mapping all modalities in a shared latent space, is another kind of method. 
Data augmentation techniques and distillation frameworks are widely applied to improve the robustness of segmentation and multiple-sensor object detection~\cite{maheshwari2023missing, li2022modality} in test time.
However, Few of the past works in RGB-T object detection consider the unexpected modality imbalance in the testing phase, which leads to an OOD issue. In this paper, inspired by many effective attempts in other fields, we propose the base-and-auxiliary architecture for robust RGB-T object detection in practice.

%% file: 3_method.tex
\section{Method}

We expect the detector to exhibit the same response to degraded imbalanced data as it does to balanced data. Data augmentation is a typical technique, which achieves this by implicitly aligning data of varying quality to the same label. However, it may disturb the convergence of training. To address this issue, we propose an architecture for training called the base-and-auxiliary detector group. It imposes a direct constraint on the detector’s output logits, enhancing the detector’s robustness by ensuring consistent responses to both RGB-T data and its imbalanced variants. The remainders of this section give a detailed presentation of our method.

\subsection{Framework}
\label{sec:ov}
We first state and model the problem. The response of the RGB-T detector is expected to depend only on the semantic information, regardless of the balance between modalities. Assuming that the RGB image is degraded, $\gamma$ and $\hat{\gamma}$ are responses of the model for two statuses of a sample:
\begin{equation}
\begin{aligned}
    \gamma &= \mathscr{F}(x), \ x=\{x_{rgb},x_{tir}\}, \\
    \hat{\gamma} &= \mathscr{F}(\hat{x}), \ \hat{x}=\{\hat{x}_{rgb},x_{tir}\},\\
\end{aligned}
\end{equation}
where $\mathscr{F}(\cdot)$ is the detector, $\hat{x}_{rgb}$ means the degraded RGB image. We can give a simple constraint with the Euclidean distance for this demand:
\begin{equation}
\label{eq:cst1}
    \min_{\mathscr{F}}\| \gamma-\hat{\gamma} \|^{2}~.
\end{equation}

However, based on the recent literature~\cite{yolox}, the model updated with momentum from its previous states tends to have higher robustness. The typical update approach is the exponential moving average~(EMA), which can be represented as:
\begin{equation}
\label{eq:ema}
\begin{aligned}
    \mathscr{F}_{ema}^{0} &= \mathscr{F}~, \\
    \mathscr{F}_{ema}^{i} \longleftarrow \alpha \cdot \mathscr{F} &+ (1-\alpha) \cdot \mathscr{F}_{ema}^{i-1}~,
\end{aligned}
\end{equation}
where $\alpha$ is the weight of the update and $i$ means the times of iteration. 

Hence, we separately deploy two unshared detectors as the base and the auxiliary detector. The base detector is an unlearned network updated by Eq.~(\ref{eq:ema}) from the auxiliary detector and acts as the source of supervision for the auxiliary one. The constraint Eq.~(\ref{eq:cst1}) can be rewritten as:
\begin{equation}
\label{eq:cst2}
    \min_{\mathscr{F}}\| \mathscr{F}_{ema}(x) - \mathscr{F}(\hat{x}) \|^{2}~.
\end{equation}

The base detector~(\ie~$\mathscr{F}_{ema}$) combines the older versions of the auxiliary one with momentum and maintains the proper response to the balanced data. By constraining the consistency between them, the modality-invariant presentation is enhanced in the auxiliary detector and propagated to the base model by updating parameters.

The implementation of the framework is illustrated in Figure~\ref{Fig:method}. The auxiliary detector has the same architecture as the base one. It is a two-stream modal quality-aware RGB-T object detector equipped with our proposed quality-aware modality interaction module. This module adapts the detector to the RGB-T data with imbalanced quality, which will be introduced in Sec~\ref{sec:int}. The auxiliary detector is supervised by the annotated RGB-T image pairs. In this stage, we conduct the pseudo degradation~(presented in Sec~\ref{sec:data_aug}) to generate imbalanced samples for learning. The parameters of the base detector are not learned directly and are updated through the EMA manner from the auxiliary detector instead. The auxiliary detector is supervised by the annotation and also receives consistency constraints from the base detector, which ensures a credible response to the degraded samples.

Considering that the pseudo degradation is not conducted on every training sample, the remainder is the additional advantage of our method. For those unchanged samples, the consistency constraint is a regularization mechanism. Each step of an update is small, making the base detector robust enough. The regularization from the base detector prevents the auxiliary one from overfitting abnormally.

\begin{figure}[!t]
    \flushleft
    \includegraphics[width=\linewidth]{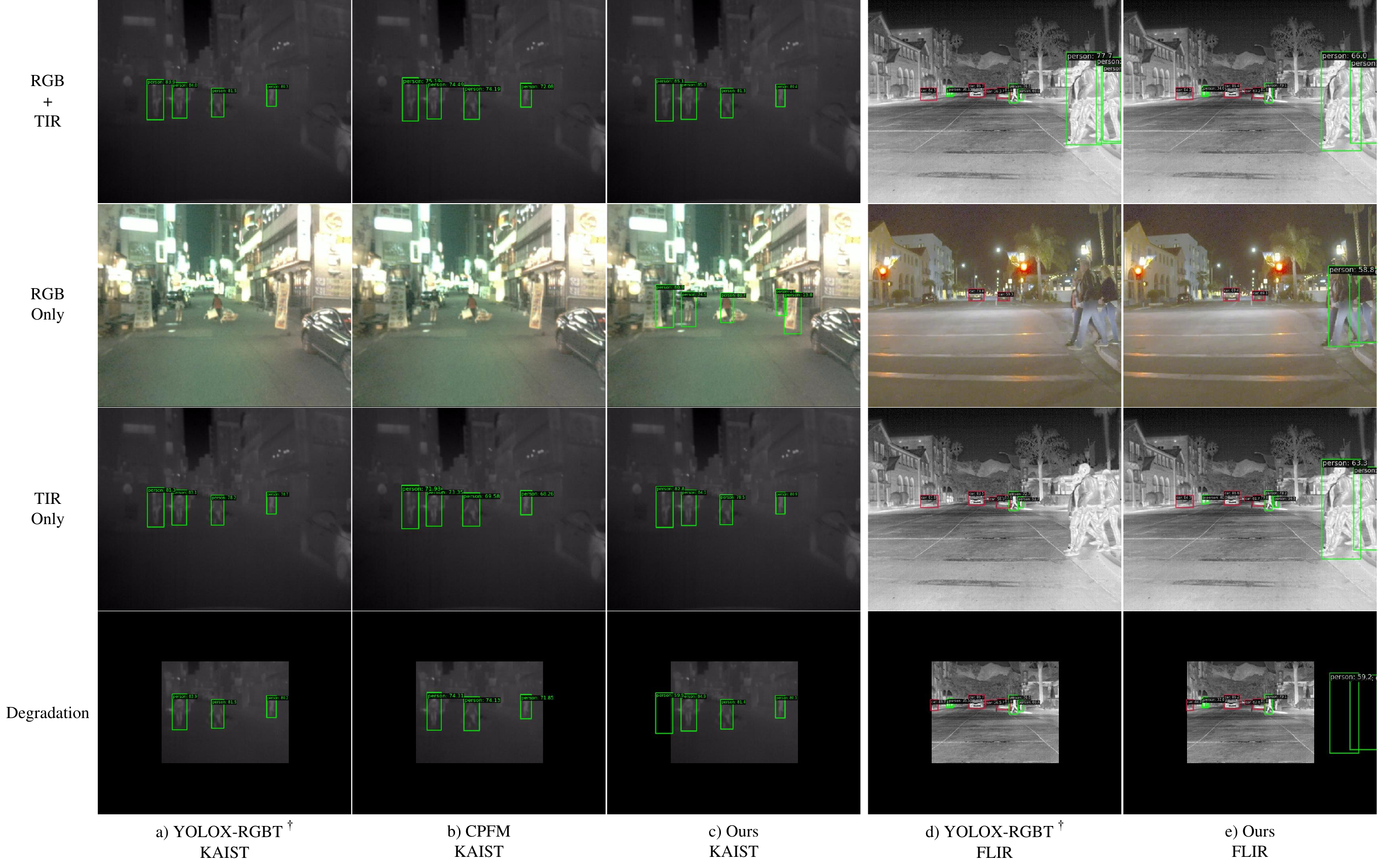}
    \caption{
    Qualitative comparison on KAIST and FLIR benchmark, including a reproduced vanilla YOLOX-RGBT, CPFM~\cite{tian2023cross} and ours in different conditions. 
    All these detectors have perfect predictions with both modalities. However, our method outperforms other detectors when meeting the imbalanced data on a local or global scale.
}
    \label{fig:img_cmp}
    \vspace{-10pt}
\end{figure}

\subsection{Quality-aware Modality Interaction Module}
\label{sec:int}
Before employing our proposed training framework, it is necessary to build a detector template. The detector should be adaptive to either balanced or imbalanced image pairs. To avoid the noise in the degraded modality, each modality should be reweighted according to their quality at different locations. We introduce our proposed modality interaction module to address the problem. Detailed network architecture is shown in Figure~\ref{Fig:method}. 
In a two-stream network, we design a concise interaction module to re-weight each modality by its quality in various areas for better fusion. 
 
The backbone has two parallel streams to tackle the RGB-T inputs $x=\{x_{rgb},x_{tir}\}$. The primary structure of each stream is the CSPDarknet~\cite{yolov4}. And the interaction begins after the $3_{rd}$ stage. The features with \textit{C} channels from \textit{k}-th stage $X_m^{(k)}\in \mathbb{R}^{N\times C\times H\times W},k\in\{3,4,5\},m\in\{rgb,tir\}$ are fed into a convolution layer activated by a Sigmoid function to predict the weight mask of the feature $M_m^{(k)}\in (0,1)^{N\times 1\times H\times W}$, where \textit{N} is the batch size and $H \times W$ is the shape of feature map. As shown in Figure~\ref{Fig:method}, the low-quality area in the degraded modality is marked with lower scores~(\ie the blue color) by predicted masks.
Then the features scaled by the mask are added to the other modality. The procedure can be formulated as:
\begin{equation}
\begin{aligned}
    M_{rgb}^{(k)} &= \Psi_{RGB}^{(k)}(X_{rgb}^{(k)}),\\
    M_{tir}^{(k)} &= \Psi_{TIR}^{(k)}(X_{tir}^{(k)}),\\
    \hat{X}_{rgb}^{(k)} &= X_{rgb}^{(k)} + M_{tir}^{(k)}\cdot X_{tir}^{(k)},\\
    \hat{X}_{tir}^{(k)} &= X_{tir}^{(k)} + M_{rgb}^{(k)}\cdot X_{rgb}^{(k)}
\end{aligned}
\end{equation}
where $\Psi$ is the convolution layer activated by the Sigmoid function and ($\cdot$) means the element-wise product. The features of a modality are weighted by its quality at a certain location, which is an attention mechanism. This proposed backbone is appropriate for the imbalanced data.

The extracted multi-stage fusion features are generated by the concatenation and convolution operation as shown in Figure~\ref{Fig:method}. Then the multi-stage fusion features are fed into the neck, which is a modified feature pyramid network used to improve the multi-scale ability of the detector. The heads then predict the object results. In general, the configurations of the neck and head are the same as those in YOLOX~\cite{yolox}.

\begin{table}[!t]
\caption{Comparison with other methods on KAIST benchmark  with three different conditions~\ie~both modalities (R+T), RGB only (R), and TIR only (T).}
\label{tab:kasit}
\centering
\footnotesize
\begin{tabular}{cccccc}
\toprule
\multirow{2}{*}{Methods} & \multirow{2}{*}{Params} & \multirow{2}{*}{Mod.} & \multicolumn{3}{c}{LAMR $\downarrow$} \\ \cmidrule(lr){4-6}
& & & All & Day & Night \\ \midrule
CIAN~\cite{cian} & - & R+T & 14.12 & 14.77 & 11.13 \\ 
AR-CNN~\cite{zhang2019weakly} & - & R+T & 9.34 & 9.94 & 8.38 \\
UGCML~\cite{kim2021uncertainty} & - & R+T & 7.89 & 8.18 & 6.96 \\
CMPD~\cite{li2022confidence} & 0.672G & R+T & 8.16 & 8.77 & 7.31 \\
CPFM~\cite{tian2023cross} & 196.02M & R+T & \underline{\textbf{6.62}} & \underline{\textbf{7.09}} & 5.61 \\ 
YOLOX-RGBT${\ }^{\dagger}$ & 51.57M & R+T & 15.52 & 19.41 & 8.7 \\
TINet~\cite{zhang2023illumination} & 97M & R+T & 9.15 & 10.25 & 7,48 \\
Ours & 72.18M & R+T & 8.63 & 10.88 & \underline{\textbf{4.69}} \\ \midrule
CMPD~\cite{li2022confidence} & 0.672G & R & 53.48 & 41.55 & 77.66 \\
CPFM~\cite{tian2023cross} & 196.02M & R & 46.7 & 30.72 & 80.9 \\
YOLOX-RGBT${\ }^{\dagger}$ & 51.57M & R & 64.55 & 53.54 & 90.25 \\
Ours & 72.18M & R & \underline{\textbf{31.35}} & \underline{\textbf{24.66}} & \underline{\textbf{44.7}} \\ \midrule
CMPD~\cite{li2022confidence} & 0.672G & T & 35.92 & 40.96 & 24.35 \\
CPFM~\cite{tian2023cross} & 196.02M & T & 17.8 & \underline{\textbf{20.52}} & 11.24 \\
YOLOX-RGBT${\ }^{\dagger}$ & 51.57M & T & 23.63 & 29.79 & 10.62 \\
Ours & 72.18M & T & \underline{\textbf{16.63}} & 22.27 & \underline{\textbf{5.87}} \\
\bottomrule
\multicolumn{6}{l}{\small $\dagger$ means our reproduced vanilla detector.}
\end{tabular}
\vspace{-10pt}
\end{table}

\subsection{Pseudo Degradation Strategy}
\label{sec:data_aug}
We herein present the manner of the pseudo degradation in modalities that we apply in our base-and-auxiliary framework. In RGB-T object detection, the degradation of a modality is reflected in the low contrast of an image on a local or global scale. Low contrast usually comes with an extreme bias, \ie too bright or too dark, causing a loss of information. Though the global degradation across the whole image is not a common occurrence, we still simulate it on the whole image for lower computational overhead. Our detector, a fully convolutional network, can learn efficiently regardless of different degradation scales.

Our pseudo degradation works with a small probability, which means the image pair is highly likely to remain unchanged. When the degradation is actually conducted, there will be only one modality to be degraded, making this pair imbalanced. In degradation, we use a multiplier \textit{c} to decline the contrast and set a bias \textit{b} for the chosen modality. The procedure can be formulated as:
\begin{equation}
    \hat{x}_m = b+c \cdot x_m, m \in \{rgb,tir\}.
\end{equation}
The multiplier obeys the uniform distribution, and the bias obeys the Gaussian distribution. These are formulated as:
\begin{equation}
\label{eq:distribution}
\begin{aligned}
    c &\sim U(0, u), u\in (0,1),\\
    b &\sim N(\mu, \sigma^2)
\end{aligned}
\end{equation}
where $u,\mu$ and $\sigma$ are hyper-parameters.

\subsection{End-to-end Learning and Inference}
\noindent
\textbf{Training.} The algorithm can be optimized by end-to-end training. The auxiliary detector is supervised by the conventional ground truth annotations, as well as the additional consistency loss between itself and the base detector~(\ie Eq.~(\ref{eq:cst2})). The total loss is formulated as:
\begin{equation}
\label{eq:loss}  \mathcal{L}_{total}=\mathcal{L}_{obj}+\mathcal{L}_{reg}+\mathcal{L}_{cls}+\mathcal{L}_{consist}~,
\end{equation}
where $\mathcal{L}_{obj},\mathcal{L}_{reg}$ and $\mathcal{L}_{cls}$ are classic detection loss. The $\mathcal{L}_{obj}$ and $\mathcal{L}_{cls}$ are Binary Cross Entropy~(BCE) loss, and the $\mathcal{L}_{reg}$ is DIoU loss. The $\mathcal{L}_{consist}$ is a L2 loss described as Eq.~(\ref{eq:cst2}).

The auxiliary detector updates parameters based on the loss function $\mathcal{L}_{total}$ while the base detector is frozen in this procedure. Then the teacher detector updates its parameters by Eq.~(\ref{eq:ema}), finishing the entire iteration.

\noindent
\textbf{Testing.} The auxiliary detector will be directly dropped after training. The base one is the final model for the practical test, as shown in Figure~\ref{Fig:method}.

\begin{table}[!t]
\caption{Comparison with other methods on FLIR benchmark.}
\label{tab:flir}
\centering
\footnotesize
\setlength{\tabcolsep}{0.015\linewidth}
\begin{tabular}{ccccccc}
\toprule
\multirow{2}{*}{Methods} & \multirow{2}{*}{Params} & \multirow{2}{*}{Mod.} & \multicolumn{4}{c}{AP50 $\uparrow$} \\ \cmidrule(lr){4-7}
 & & & mAP & Car & Bicycle & Person \\ \midrule
CMPD~\cite{li2022confidence} & 0.672G & R+T & 69.35 & 78.11 & 59.87 & 69.64 \\
CFR\_3~\cite{zhang2020multispectral_flir} & - & R+T & 72.39 & 84.91 & 57.77 & 74.49 \\
GAFF~\cite{gaff} & - & R+T & 72.9 & - & - & - \\
YOLOFus.~\cite{yolofusion} & - & R+T & 76.6 & - & - & - \\
YOLOX-RGBT${\ }^{\dagger}$ & 51.57M & R+T & 75.5 & 87.9 & 54.5 & 84 \\
IGT~\cite{chen2023igt} & $>>$137M & R+T & \underline{\textbf{85}} & - & - & - \\
CrossFormer\cite{lee2024crossformer} & $>$120M & R+T & 79.3 & - & - & - \\
Ours & 72.18M & R+T & 78.8 & \underline{\textbf{89.3}} & \underline{\textbf{62}} & \underline{\textbf{85}} \\ \midrule
YOLOX-RGBT${\ }^{\dagger}$ & 51.57M& R & 40.1 & 62.1 & 24.3 & 33.7 \\
Ours & 72.18M & R & \underline{\textbf{52.5}} & \underline{\textbf{74.4}} & \underline{\textbf{33.8}} & \underline{\textbf{49.4}} \\ \midrule
YOLOX-RGBT${\ }^{\dagger}$ & 51.57M & T & 43.3 & 57.2 & 24.9 & 47.8 \\
Ours & 72.18M & T & \underline{\textbf{75.2}} & \underline{\textbf{87.4}} & \underline{\textbf{55.9}} & \underline{\textbf{82.3}} \\
\bottomrule
\multicolumn{6}{l}{\small $\dagger$ means our reproduced vanilla detector.}
\end{tabular}
\vspace{-10pt}
\end{table}

%% file: 4_experiment.tex
\section{Experiments}

\subsection{Experimental Settings}
\textbf{Datasets.}  
The KAIST~\cite{kaist} dataset is a widely used RGB-T pedestrian detection benchmark. The dataset provides paired RGB-T images with $512 \times 640$ resolution, where 7000 pairs are for training, and 2000 pairs are for testing. The original annotations of KAIST are low-quality, Liu \etal~\cite{liu2016multispectral}~provide a re-annotated version, which is applied in our experiment. The FLIR ADAS dataset\footnote{https://www.flir.com/oem/adas/adas-dataset-form/} is another multi-class RGB-T dataset. 
Zhang~\etal~\cite{zhang2020multispectral_flir} provide a registered and scrubbed subset with $512 \times 640$ resolution, where the categories with few samples are eliminated, and 
three main categories are retained, \ie~car, bicycle, and person. We utilize this benchmark to evaluate our proposed method as other works~\cite{li2022confidence, gaff}.

\noindent \textbf{Evaluation Settings.} The KAIST benchmark uses the Log-average Missing Rate~(LAMR) as the metric. The range of False Positive Per Image~(FPPI) is set to $[10^{-2},10^0]$ in log space, and the Intersection over Union~(IOU) threshold is set to 0.5 for the MR calculating. A lower value is better for LAMR. The FLIR benchmark adopts the COCO-style average precision~(AP) as the test metric, where the IOU threshold for AP is set to 0.5.

\subsection{Implementation Details}
\begin{figure}[!t]
\centering
\caption{The $\mu$ and $\sigma$ are solved by a linear scaling principles. The scaling maps the range of [-2.5, 2.5] to [0,255], where the integral probability is 0.995.
}
\includegraphics[width=0.4\linewidth]{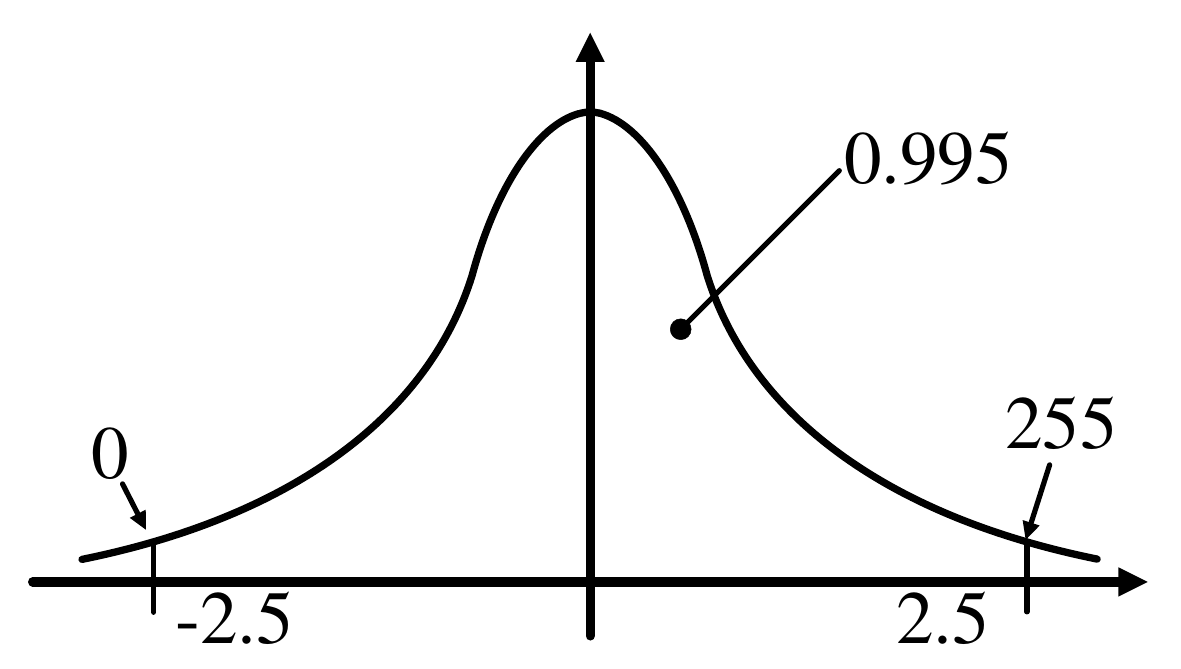}
\label{fig:gaussian}
\vspace{-10pt}
\end{figure}

The proposed method is implemented with the MMdet\footnote{https://github.com/open-mmlab/mmdetection} toolkit. In the detector, the main body of the backbone consists of two parallel CSPDarknet53s, each of which is scaled by a depth factor $s_{depth}=0.67$ and a width factor $s_{width}=0.75$ to slash the layers and the channels of each layer. The factors tend to be adopted as the medium level in the YOLO series~(like YOLOv5-M). Except for the interaction modules, the remainders of the detector are exactly the same as those of YOLOX~\cite{yolox}. In pseudo degradation, the probability of conducting is set to 0.3. The hyper-parameters $u$ in Eq.~\ref{eq:distribution} is set to $0.7$, and $\mu$ and $\sigma^2$ are set to $127.45$ and $2440$ respectively, which ensures the probability of bias $b \in [0,255]$ to be $99.5\%$. We solve these values by a linear scale of standard normal distribution, where we map the range of [-2.5, 2.5] to [0, 255], shown in Figure.\ref{fig:gaussian}.

Constrained by the previous states, the auxiliary model tends to get a smaller evolution. A too-small replacement rate of EMA thus disturbs the model convergence. We set the replacement rate in Eq.~\ref{eq:ema} to $\alpha=0.001$. We adopt the SGD optimizer throughout our experiment. The learning rates are set to $0.01$ and $0.02$ respectively for KAIST and FLIR datasets. The models are trained for 20 and 47 epochs respectively on two datasets with two RTX 3090ti GPUs.

\begin{table}[!t]
\caption{Ablation study on KAIST benchmark. We test the effectiveness of proposed interaction modules~(Int.), pseudo degradation~(P.D.), and auxiliary detector~(Aux.).}
\label{tab:ablation}
\centering
\footnotesize
\setlength{\baselineskip}{1.2\baselineskip}
\setlength{\tabcolsep}{0.02\linewidth}
\begin{tabular}{ccccccccc}
\toprule
\multirow{3}{*}{Int.} & \multirow{3}{*}{P.D.} & \multirow{3}{*}{Aux.} & \multicolumn{6}{c}{LAMR $\downarrow$} \\ \cmidrule(lr){4-9}
 & & & \multicolumn{2}{c}{R+T} & \multicolumn{2}{c}{R} & \multicolumn{2}{c}{T} \\ \cmidrule(lr){4-5}  \cmidrule(lr){6-7} \cmidrule(lr){8-9}
 & & & Day & Night & Day & Night & Day & Night \\ 
\midrule
 & & & 19.41    & 8.7   & 53.54 & 90.25 & 29.79 & 10.62 \\[0.5mm]
 & \checkmark & & 23.1  & 7.71  & 39.95 & 56.6  & 39.78 & 8.5 \\[0.5mm]
\checkmark & & & 16.93  & 7.37  & 61.08 & 93.35 & 37.68 & 11.74 \\[0.5mm]
\checkmark & \checkmark & & 19.29 & 8.61 & 41.49 & 62.45 & 32.03 & 10.16 \\[0.5mm]
\checkmark & & \checkmark   & 11.11 & 4.78  & 43.63 & 82.64 & \underline{\textbf{21.28}} & 6.91 \\[0.5mm]
\checkmark & \checkmark & \checkmark & \underline{\textbf{10.88}} & \underline{\textbf{4.69}} & \underline{\textbf{24.66}} & \underline{\textbf{44.7}}  & 22.27 & \underline{\textbf{5.87}} \\[0.5mm]

\bottomrule
\end{tabular}
\vspace{-10pt}
\end{table}

\subsection{Comparison}
Herein, we make a comparison between our method and the typical algorithms in RGB-T object detection.

\textbf{KAIST.}
We first conduct a qualitative comparison including our proposed method, the native RGB-T detector CPFM~\cite{tian2023cross}, and our reproduced RGB-T detector based on YOLOX~\cite{yolox}. As shown in Figure~\ref{fig:img_cmp}, the thermal image is proved to be important for RGB-T detectors for its all-weather advantages. Missing thermal modality on either a global or local scale disturbs the prediction of all detectors. In the second row~(RGB only), the YOLOX-RGBT and CPFM fail to detect any pedestrian. However, our proposed detector suffers a limited impact and gives the predictions with only one false positive object, proving to be more robust in different practical conditions.

The quantitative comparison of the evaluation at the KAIST benchmark is shown in Table~\ref{tab:kasit}. Most works report the LAMR tested only with the balanced data~(\ie~Both two modalities). With normal data, our detector equipped with a smaller backbone achieves competitive performance compared to other state-of-the-art methods. With imbalanced data, all methods suffer from a large decline in LAMR. Especially in the RGB-only condition, the drop in performance indicates that most RGB-T detectors rely actually on thermal images. However, our method shows the superiority to all other detectors by a large margin with fewer parameters. For example, CPFM, currently the most powerful dedicated model in balancing RGB-T data, exhibits a significant performance degradation when losing one modality. Specifically, its Log-Average Miss Rate (LAMR) drastically increases from 6.62 to 46.7 upon losing the TIR modality, corresponding to a 605\% increase. Similarly, when the RGB modality is missing, the LAMR rises to 17.8, reflecting a 169\% increase. In contrast, under similar conditions, the increases for our proposed model are 263\% and 92.7\%, respectively.

\textbf{FLIR.}
Considering most benchmarks have only the pedestrian class, we leverage the FLIR benchmark to evaluate the multi-class detection performance. A few detectors report the testing results while our proposed detector outperforms most methods with similar parameter quantities by a large margin in all classes. Though the IGT~\cite{chen2023igt} has a higher mAP, it is a transformer-based detector with many more parameters.
The comparison indicates that our detector has a favorable classification capability, even if encountering an extreme modality imbalance. These results indicate that our method has the potential to be extended to other datasets with more classes. There is also a qualitative comparison shown in Figure~\ref{fig:img_cmp}.

\subsection{Ablation Study} \label{sec:ablation}

We test the contribution of each component to the modal robustness. As shown in Table~\ref{tab:ablation}, all models are trained with balanced dataset~(\ie~the original training set), and tested with three modality modes~(Both, RGB-only and TIR-only).

\textbf{Effect of interaction modules.}
The third state in Table~\ref{tab:ablation} is the base detector, a two-branch variant of YOLOX. The concatenated features from two unshared backbones are used for further inference. We design an interaction module for adaptive feature fusion. The module works well for balanced data~(\ie~with Both modalities) but fails in imbalanced data~(\ie~second state) because it exacerbates the OOD issue for those unseen samples in the test.

\textbf{Effect of pseudo degradation.}
The pseudo degradation plays the role of data augmentation. Data augmentation is a typical solution to relieve the potential OOD issue in the test stage and is a main rival for comparison. As shown in the second state in Table~\ref{tab:ablation}, the vanilla data augmentation improves the performance in tackling those unseen imbalanced samples in the test. 

However, according to our experiments, the straight strong data augmentation disturbs the convergence of the network and leads to a performance decline for normal samples, which is shown as the "R+T" mode in the second line in Table~\ref{tab:ablation}.

\begin{figure}[!t]
\centering
\caption{Vanilla modal augmentation disturbs the convergence of the model. 
Our method effectively improves the robustness of training on the KAIST benchmark.
}
\includegraphics[width=0.8\linewidth]{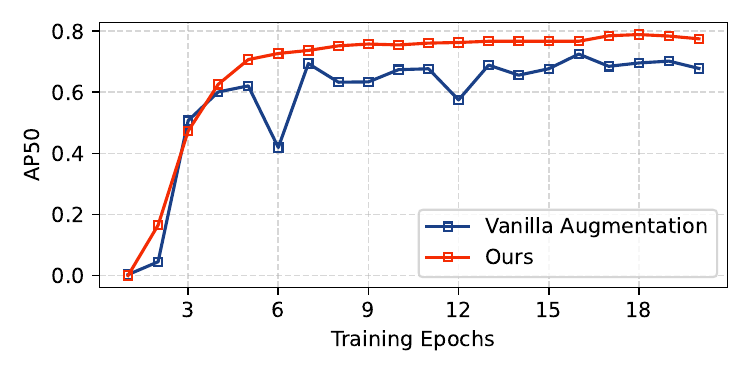}
\label{fig:converg}
\vspace{-10pt}
\end{figure}

\begin{table}[!]
\centering
\footnotesize
\caption{The LAMR comparison between the base and the auxiliary detector. The base detector performs better even though the auxiliary one is regularized.}
\label{tab:base_aux}
\begin{tabular}{ccccc}
\toprule
\multirow{2}{*}{Final detector} & \multicolumn{3}{c}{LAMR $\downarrow$}    & \multirow{2}{*}{Mod.}  \\ \cmidrule(lr){2-4} 
 & All  & Day   & Night & \\ \midrule
\multirow{3}{*}{Base detector}  & \textbf{8.63}  & \textbf{10.88} & \textbf{4.69}  & R+T \\ 
                                & 31.35 & 24.66 & 44.7  & R \\
                                & 16.63 & 22.27 & 5.87  & T \\ \midrule
\multirow{3}{*}{Auxiliary detector} & \underline{12.68} & \underline{16.34} & \underline{6.04}  & R+T \\
                                    & 40.33 & 30.79 & 59.57 & R \\
                                    & 21.65 & 28.35 & 8.01  & T \\
\bottomrule                             
\end{tabular}
\vspace{-10pt}
\end{table}

{
\setlength{\tabcolsep}{4.5pt} 
\begin{table}[t]
\footnotesize
\centering
\caption{Robustness to Gaussian noise on KAIST. The underlined results demonstrate the robustness of our training architecture combined with the pseudo-degradation strategy against high-intensity Gaussian noise.}
\label{tab:gaussian_noise}
\begin{tabular}{lclcccc}
\toprule
\multirow{2}{*}{Models} & \multirow{2}{*}{\makecell{Training \\ Aug.}} & \multirow{2}{*}{\makecell{Testing\\Noise}} & \multirow{2}{*}{Add-to} & \multicolumn{3}{c}{LAMR $\downarrow$} \\ \cmidrule(lr){5-7}
     &     &    &         & All   & Day   & Night \\
\midrule
YOLOX-RGBT${\ }^{\dagger}$    & -   & $\sigma=0$    & -   & 15.52 & 19.41 & 8.70 \\
YOLOX-RGBT${\ }^{\dagger}$    & -   & $\sigma=20$   & RGB & 16.77 & 20.41 & 10.42 \\
YOLOX-RGBT${\ }^{\dagger}$    & -   & $\sigma=20$   & TIR & 25.15 & 21.96 & 23.71 \\ \midrule
Ours          & Pse-D   & $\sigma=0$    & -   & \underline{8.63}  & 10.88 & 4.69  \\
Ours          & Pse-D   & $\sigma=20$   & RGB & \underline{8.86}  & 11.18 & 4.66  \\
Ours          & Pse-D   & $\sigma=20$   & TIR & \underline{9.56}  & 10.21 & 7.06  \\ \midrule
Ours-Gaussian & 20  & $\sigma=0$    & -   & \textbf{8.61}  & 11.34 & 3.80  \\
Ours-Gaussian & 20  & $\sigma=20$   & RGB & \textbf{8.68}  & 11.32 & 4.02  \\
Ours-Gaussian & 20  & $\sigma=20$   & TIR & \textbf{9.43}  & 11.96 & 4.82  \\
\bottomrule
\end{tabular}
\end{table}
}

\begin{figure*}[t]
    \centering
    \includegraphics[width=0.28\linewidth]{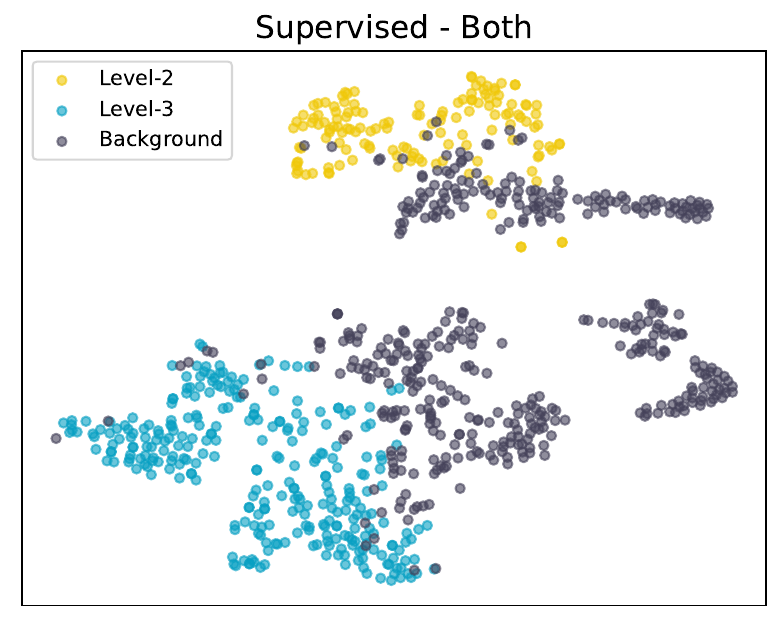}
    \includegraphics[width=0.28\linewidth]{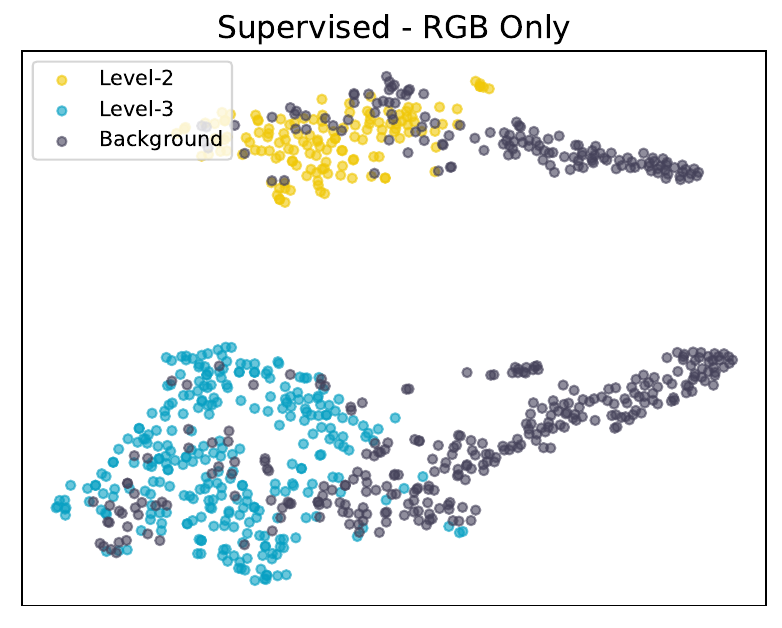}
    \includegraphics[width=0.28\linewidth]{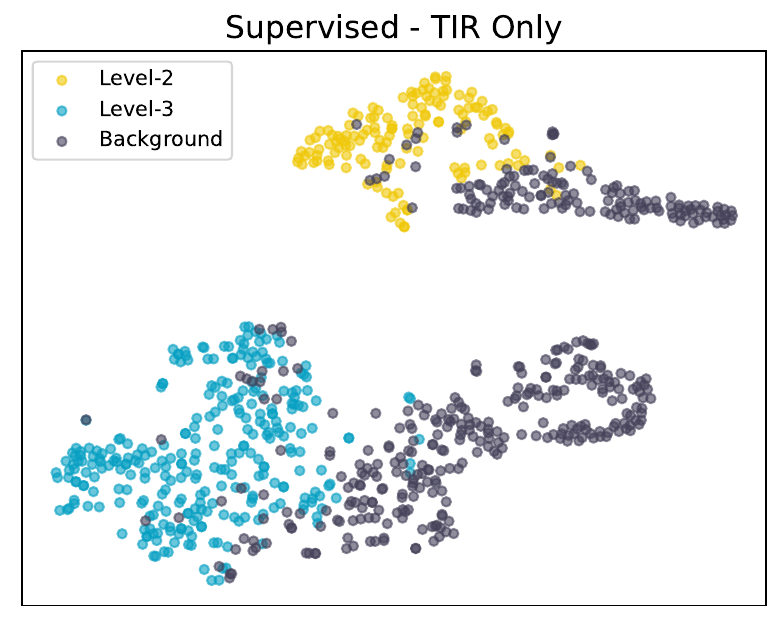}
    
    \includegraphics[width=0.28\linewidth]{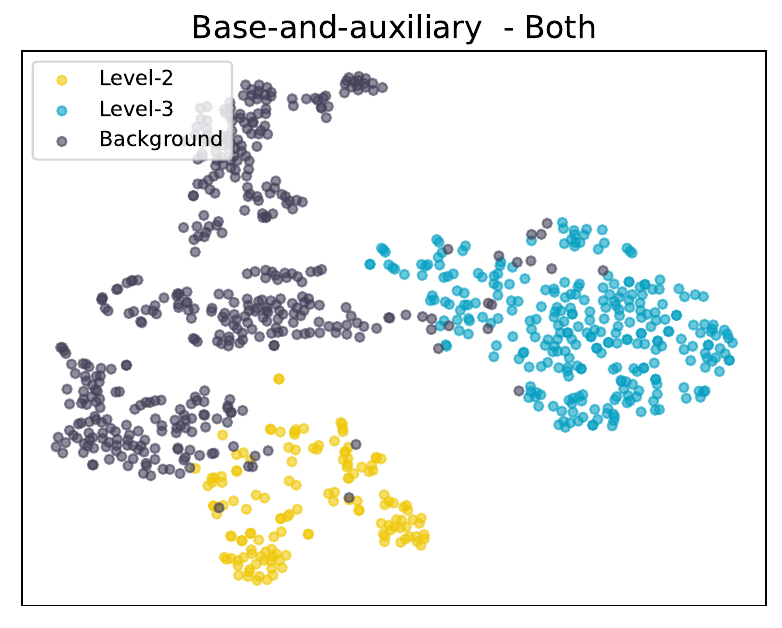}
    \includegraphics[width=0.28\linewidth]{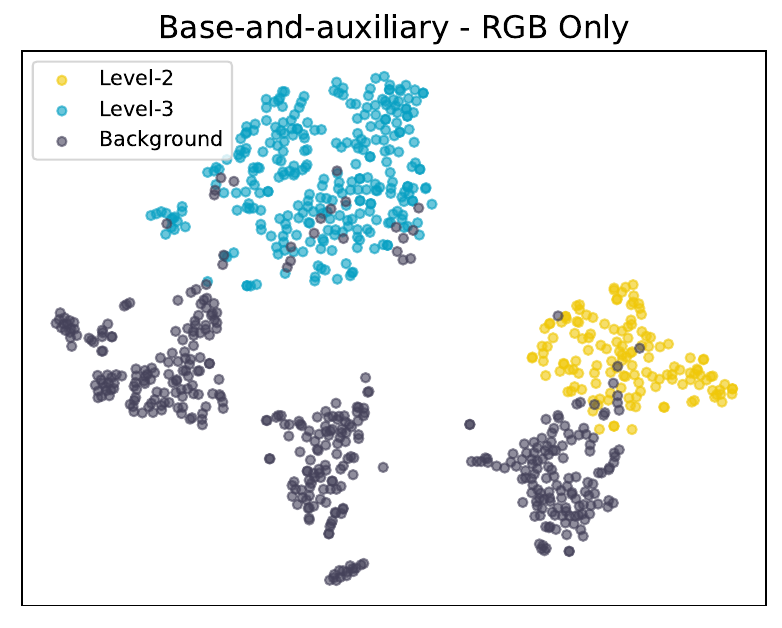}
    \includegraphics[width=0.28\linewidth]{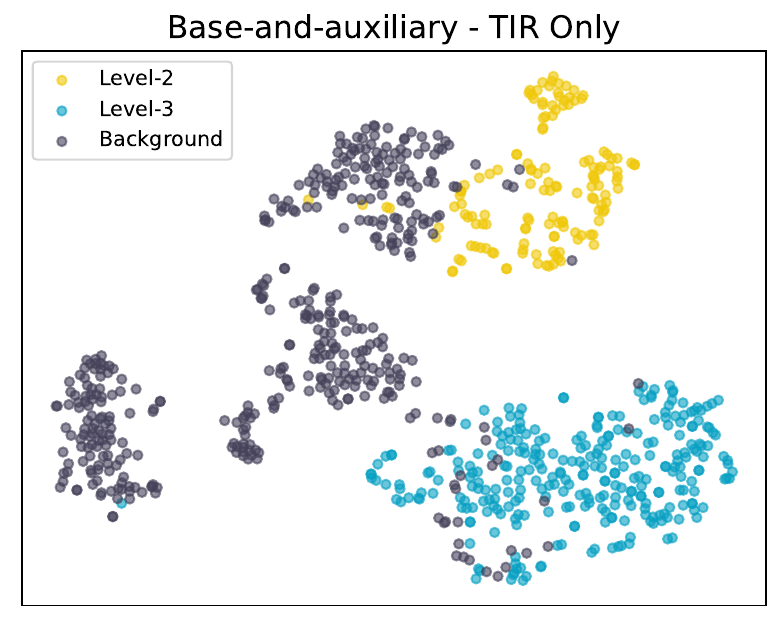}
    \caption{The visualized features of positive samples in the feature map. 
    The yellow and blue points mean the positive samples from different levels in FPN. The dark ones are features from background areas. The base detector supervised only with the annotations fails to extract the discriminative features for the imbalanced data. The features from the base detector in our proposed framework have a higher intra-class compactness.
    }
    \label{fig:tsne}
    \vspace{-10pt}
\end{figure*}

\begin{table*}[!t]
\caption{Proposed base-and-auxiliary training framework promotes the robustness of other RGB-T object detectors against contrast degradation on the KAIST benchmark. The Sup. and Aux. mean the supervised training and our base-and-auxiliary framework respectively. The "+Aux." methods (ours) are always better than their baselines. The red and green records show the best improvement by 55\%.}
\label{tab:contrast}
\centering
\scriptsize
\setlength{\tabcolsep}{0.007\linewidth}
\begin{tabular}{cccccccccccccccc}
\toprule
\multirow{4}{*}{Methods} & \multicolumn{15}{c}{LAMR $\downarrow$} \\ \cmidrule(lr){2-16}
 & \multicolumn{3}{c}{\multirow{2}{*}{Origin-Test}} & \multicolumn{6}{c}{w/o RGB} & \multicolumn{6}{c}{w/o TIR} \\ \cmidrule(lr){5-10} \cmidrule(lr){11-16}
 & & & & \multicolumn{3}{c}{0.5} & \multicolumn{3}{c}{0} & \multicolumn{3}{c}{0.5} & \multicolumn{3}{c}{0} \\ \cmidrule(lr){2-4} \cmidrule(lr){5-7} \cmidrule(lr){8-10} \cmidrule(lr){11-13} \cmidrule(lr){14-16}    
 & All & Day & Night & All & Day & Night & All & Day & Night & All & Day & Night & All & Day & Night \\ \cmidrule(lr){1-1} \cmidrule(lr){2-4} \cmidrule(lr){5-7} \cmidrule(lr){8-10} \cmidrule(lr){11-13} \cmidrule(lr){14-16} 
Base+Sup. & 13.46 & 16.93 & 7.37 & 14.15 & 17.81 & 7.9 & 28.96 & 37.68 & 11.74 & 14.31 & 17.05 & 8.89 & \red{\textbf{70.72}} & 61.08 & 93.35 \\
Base+Aux. & \textbf{8.63} & \textbf{10.88} & \textbf{4.69} & \textbf{8.4} & \textbf{10.65} & \textbf{4.47} & \textbf{16.63} & \textbf{22.27} & \textbf{5.87} & \textbf{8.8} & \textbf{11.03} & \textbf{4.71} & \green{\textbf{31.35}} & \textbf{24.66} & \textbf{44.7} \\ \cmidrule(lr){1-1} \cmidrule(lr){2-4} \cmidrule(lr){5-7} \cmidrule(lr){8-10} \cmidrule(lr){11-13} \cmidrule(lr){14-16}
VFNet~\cite{varifocalnet} & 28.2 & 33.37 & 17.09 & 30.89 & 38.56 & 15.49 & 48.89 & 57.46 & 29.92 & 38.35 & 38.46 & 35.8 & 98.3 & 97.56 & 100 \\
VFNet+Aux. & 22.85 & 28 & 12.7 & 24.37 & 30.81 & 11.98 & 39.77 & 48.27 & 22.05 & 24.85 & 28.13 & 18.03 & 60.64 & 51.28 & 80.93 \\ \cmidrule(lr){1-1} \cmidrule(lr){2-4} \cmidrule(lr){5-7} \cmidrule(lr){8-10} \cmidrule(lr){11-13} \cmidrule(lr){14-16}
SSD~\cite{ssd} & 48.42 & 56.31 & 27.4  & 48.42 & 56.31 & 27.4  & 48.42 & 56.31 & 27.4  & 43.43 & 50.88 & 24.96 & 100   & 100   & 100   \\
SSD+Aux. & 31.39 & 37.48 & 18.28 & 29.87 & 36.64 & 15.97 & 34.73 & 43.08 & 16.35 & 34.59 & 38.63 & 26.01 & 66.77 & 59.92 & 79.92 \\ 
\bottomrule
\end{tabular}
\vspace{-10pt}
\end{table*}

\textbf{Effect of Base-and-auxiliary framework.}
As shown in Table~\ref{tab:ablation}, the benefits of our proposed Base-and-auxiliary framework can be summarized as two points.

1)~Modality consistency. By introducing the proposed framework, the auxiliary detector fed with degraded samples is enforced to fit the base detector taking the normal data, which regularizes the response of the detector encountering the imbalanced RGB-T samples. 

2)~Better convergence. The updating with momentum overcomes the side effect of pseudo degradation. As shown in the 5th and 6th states in Table~\ref{tab:ablation}, the LAMR gets a comprehensive decrease in all modality settings. The modal degradation enhances the robustness of the base detector in dealing with the imbalanced data. The comparison in Figure~\ref{fig:converg} points out the performance gap between the vanilla pseudo degradation and our integrated framework.

\textbf{Comparison between the base and auxiliary detectors.}
Regularized by the base detector, the auxiliary detector obviously gets better robustness against the modality imbalance. It outperforms most detectors in extreme conditions, as shown by the comparison between Table~\ref{tab:kasit} and Table~\ref{tab:base_aux}.

And the auxiliary detector~(shown in Table~\ref{tab:base_aux}) is even better than a vanilla base detector trained in a supervised manner~(\ie~the 4th state in Table~\ref{tab:ablation}). Due to the poor annotation quality in the existing RGB-T datasets, the base detector can easily benefit from the smoother parameter update. The base detector thus outperforms the auxiliary detector in Table~\ref{tab:base_aux}.

\textbf{Features Visualization.}
To qualitatively explain our method, we visualize the FPN features of positive samples by t-SNE~\cite{tsne}, as shown in Figure~\ref{fig:tsne}. 
We compare our approach to a baseline detector trained conventionally (3rd state in Table~\ref{tab:ablation}) to highlight the improvements. 
We observe that features from our method exhibit higher intra-class compactness and maintain clear decision boundaries, particularly for imbalanced samples. 
This makes it easier for the detector to distinguish challenging objects from the background.

\textbf{Robustness to imbalanced noise.} 
Modality degradation may not be limited to contrast decay alone; thus, we have also considered other forms of degradation, such as noise. To evaluate the robustness of our detector, we introduced high-intensity Gaussian noise during testing, and the corresponding results are presented in Table~\ref{tab:gaussian_noise}. The underlined results demonstrate the robustness of our training architecture combined with the pseudo degradation strategy against high-intensity Gaussian noise. Furthermore, targeted degradation modeling, specifically for Gaussian noise, can further enhance the robustness of the detector, as highlighted by the boldface results.

\textbf{Replacement rate $\alpha$.}
The replacement rate $\alpha$ in Eq.~\ref{eq:ema} is a hyper-parameter, which controls the update speed from the auxiliary detector. A small $\alpha$ brings the smoother but much slower convergence of training. A larger $\alpha$ leads to an unstable training and the model gets a lower precision. We try the $\alpha$ from 1e-4 to 1e-1 and choose $\alpha=1e-3$ in our experiments.

\subsection{Extended Application}

The proposed base-and-auxiliary framework can be introduced not only to our two-stream detector but also to other prospective detectors. Few RGB-T detectors are open source, making it hard to evaluate the precision improvement. We thus implement several primitive two-stream RGB-T detectors, which are modified from standard RGB detectors by mirroring the structure of their backbone. We re-train these detectors on the KAIST dataset and test them elaborately in the typical modality degradation~(\ie~contrast degradation). The results are shown in Table~\ref{tab:contrast}. The Base detector in these tables is our proposed quality-aware detector. The Sup. and Aux. indicate the manners of training,~\ie~the supervised training and our base-and-auxiliary framework.

As shown in Table~\ref{tab:contrast}, the contrast of a certain modality is decreased by 0.5 and 0 times respectively. All the tested detectors are temporarily implemented, not the native ones for RGB-T object detection, exhibiting poor performance. 
By introducing our proposed architecture, the two classic detectors get large improvements in robustness.
These experiments show the potential of our method in improving the robustness of detectors against harsh conditions in practice.

%% file: 5_conclu.tex
\section{Conclusion}

In this paper, we propose an architecture for robust RGB-T object detection, particularly addressing unseen imbalanced samples under harsh conditions. Specifically, a simple modality interaction module is introduced to construct a basic two-stream RGB-T detector, enhancing its ability to manage imbalanced inputs. Additionally, modality pseudo degradation is employed during training to simulate and address data imbalance. Finally, the proposed base-and-auxiliary architecture is utilized to alleviate convergence difficulties arising from significant training data shifts. Nevertheless, within this practically oriented work, there remains room for further improvement regarding robustness against modality information loss. Additionally, the current model size exceeds the acceptable range typically required for industrial deployment. Therefore, considerable research potential exists in model lightweighting, which we regard as an important direction for our future work.